\documentclass[letterpaper, 10 pt, conference]{ieeeconf}

\IEEEoverridecommandlockouts    

\usepackage{color}
\usepackage{amsmath}
\usepackage{prettyref}
\usepackage{subcaption}
\usepackage[pdftex]{graphicx}
\usepackage{algorithm}
\usepackage{verbatim}
\usepackage{siunitx}
\usepackage{graphicx}
\usepackage{algorithm}
\usepackage[noend]{algpseudocode}
\algnewcommand{\Inputs}[1]{%
  \State \textbf{Inputs:}
  \Statex \hspace*{\algorithmicindent}\parbox[t]{.8\linewidth}{\raggedright #1}
}

\graphicspath{{images/}}
\title{\LARGE \bf
A Single-Planner Approach to Multi-Modal Humanoid Mobility
}

\author{
    Andrew Dornbush$^{1}$,
    Karthik Vijayakumar$^{1}$,
    Sameer Bardapurkar$^{1}$,
    Fahad Islam$^{1}$,
    Maxim Likhachev$^{1}$
    \thanks{$^{1}$Robotics Institute, Carnegie Mellon University, Pittsburgh, PA}
}

\begin{document}

\maketitle

\begin{abstract}

In this work, we present an approach to planning for humanoid mobility. Humanoid
mobility is a challenging problem, as the configuration space for a humanoid
robot is intractably large, especially if the robot is capable of performing
many types of locomotion. For example, a humanoid robot may be able to perform
such tasks as bipedal walking, crawling, and climbing. Our approach is to plan
for all these tasks within a single search process. This allows the search to
reason about all the capabilities of the robot at any point, and to derive the
complete solution such that the plan is guaranteed to be feasible. A key
observation is that we often can roughly decompose a mobility task into a
sequence of smaller tasks, and focus planning efforts to reason over much
smaller search spaces. To this end, we leverage the results of a recently
developed framework for planning with adaptive dimensionality, and incorporate
the capabilities of available controllers directly into the planning process.
The resulting planner can also be run in an interleaved fashion alongside
execution so that time spent idle is much reduced.

\end{abstract}

%
%

\section{Introduction}

Recent years have shown much interest in developing robust humanoid robots that
can operate in environments that are often unstructured, cluttered, and
unpredictable compared to controlled industrial settings. Furthermore, the
structure that does exist is intended primarily for people, and not designed
with the robots in mind. Structures such as staircases, ladders, railings,
complement people's mobility. This leads us to design humanoids so that they
possess capabilities similar to people such as the ability to walk, climb, and
use surfaces such as handrails for support.

The need for all these capabilities provides a number of challenge problems for
motion planning. The most pronounced problem is the inherent high dimensionality
of the robot's configuration space. To guarantee that a plan safely and
efficiently accomplishes a given task may require reasoning about all of the
joints of the robot and the relationship between the robot and the various
objects in its environment. These constraints are expensive to evaluate.

\begin{figure}
    \centering
    \begin{subfigure}{0.48\linewidth}
        \includegraphics[width=\linewidth]{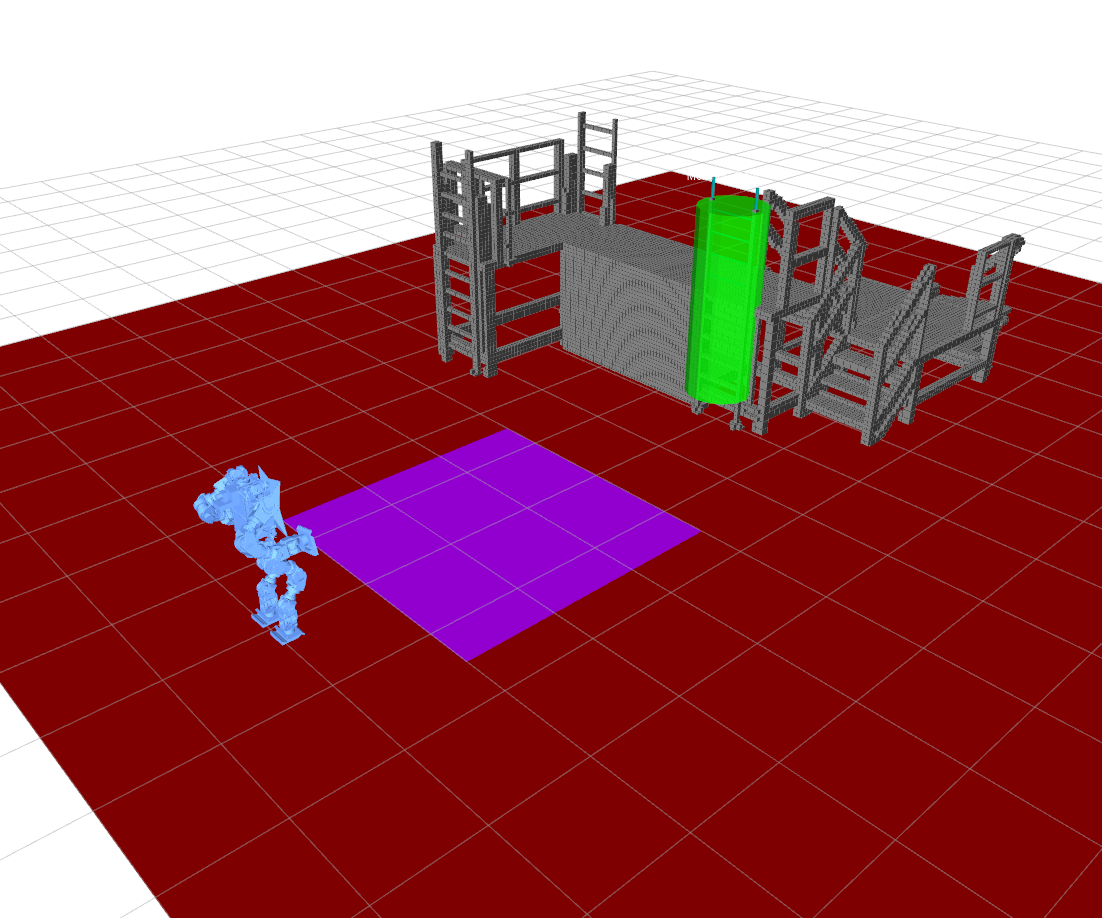}
        \caption{}
        \label{fig:test_env_1}
    \end{subfigure}
    \begin{subfigure}{0.48\linewidth}
        \includegraphics[width=\linewidth]{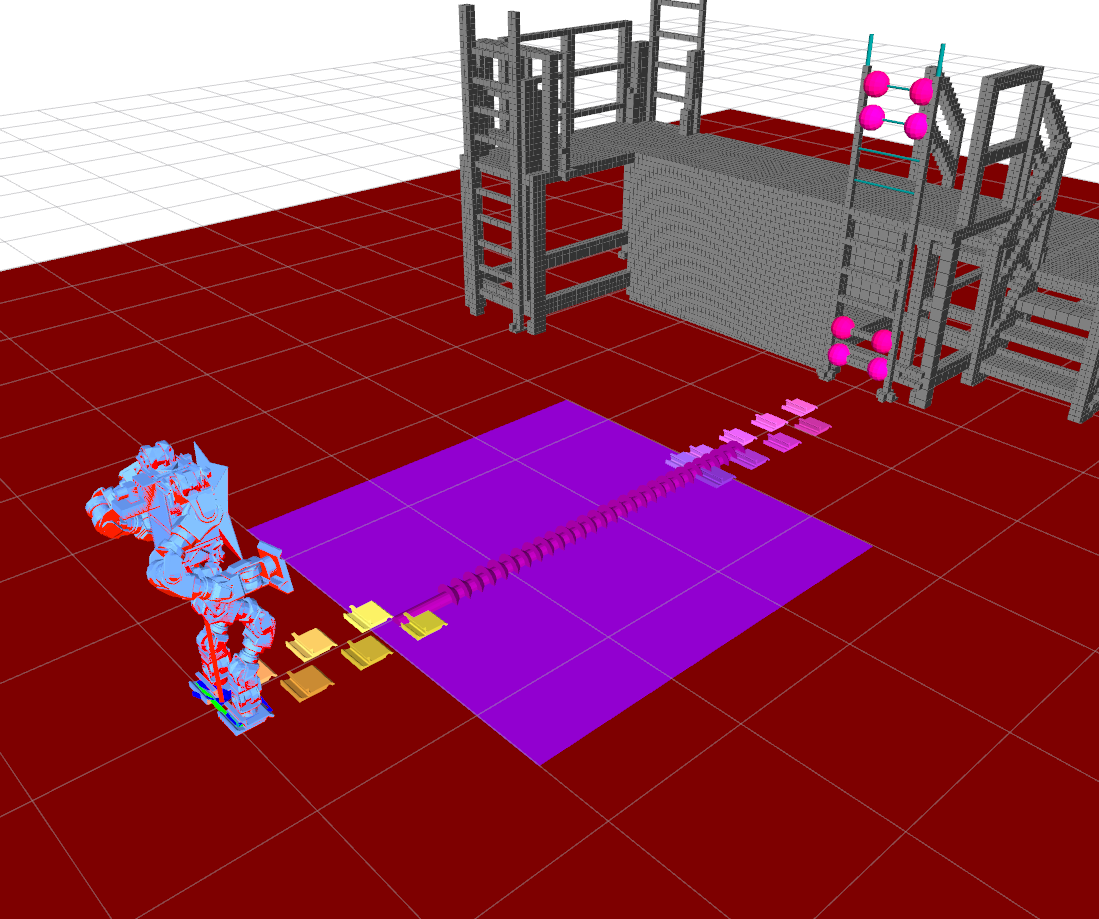}
        \caption{}
        \label{fig:test_env_1_plan}
    \end{subfigure}
    \begin{subfigure}{0.48\linewidth}
        \includegraphics[width=\linewidth]{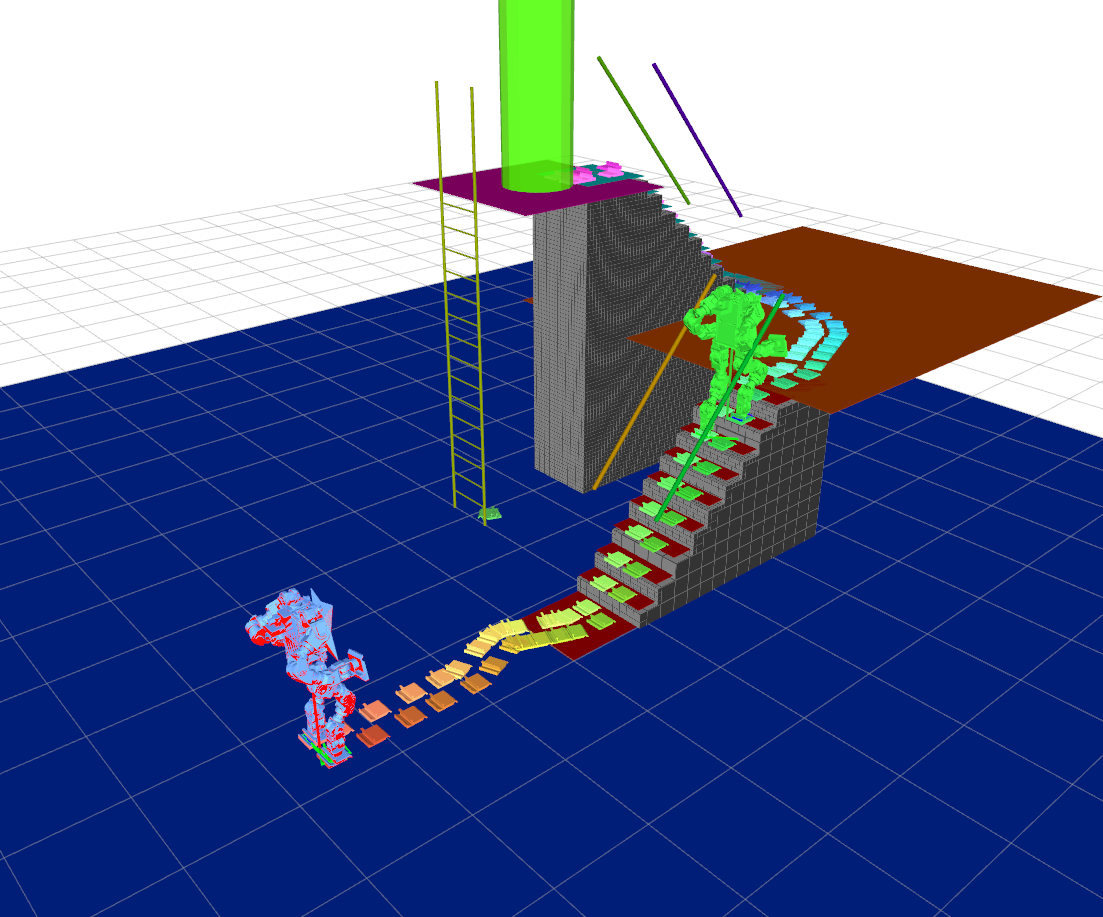}
        \caption{}
        \label{fig:test_env_2}
    \end{subfigure}
    \begin{subfigure}{0.48\linewidth}
        \includegraphics[width=\linewidth]{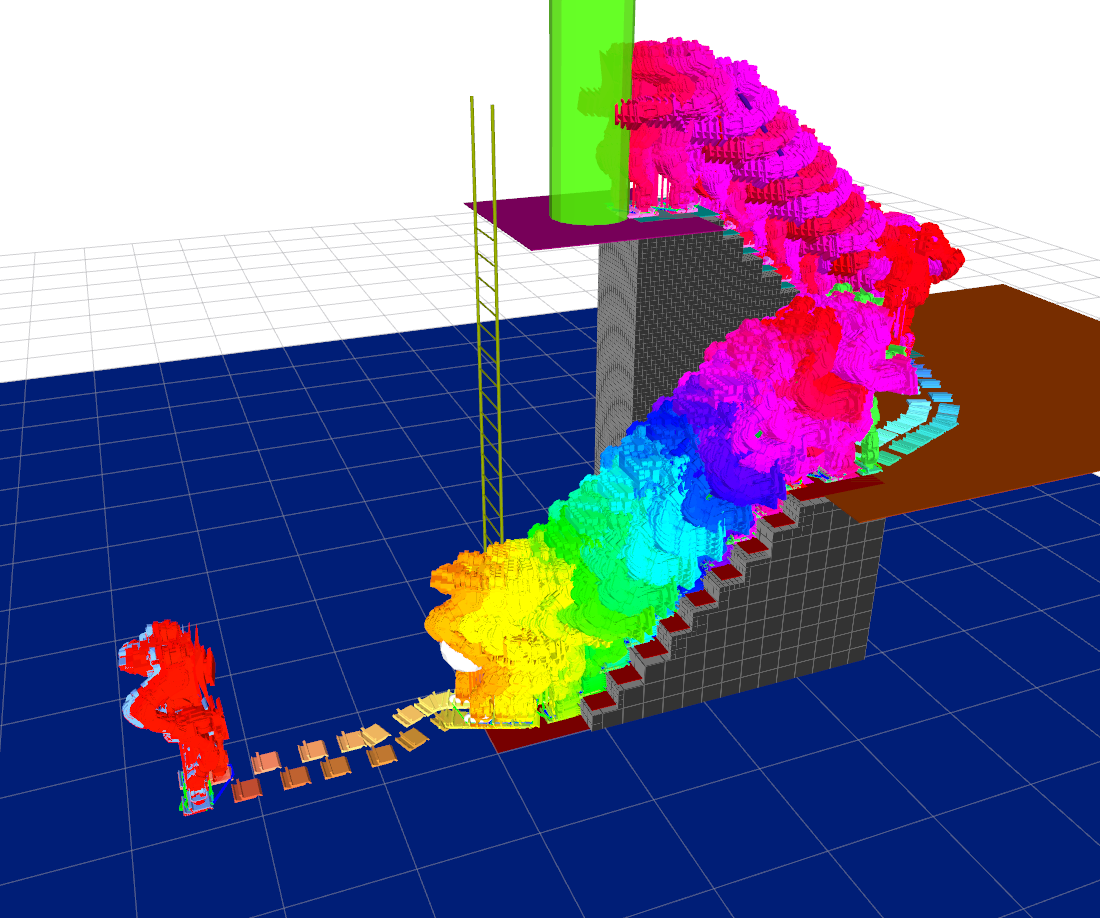}
        \caption{}
        \label{fig:test_env_2_plan}
    \end{subfigure}
    \caption{Example Environments and Plans - Hazardous "crawl-only" zones are
    depicted as purple regions in the top two figures}
    \label{fig:examples}
\end{figure}

Luckily, a complicated mobility task can often be broken down into a sequence of
smaller tasks. For example, a task for a robot to move from one end of a
facility to the other might include traversing large areas by walking, climbing
staircases or ladders and, in situations where the environments is hazardous,
crawling under fallen structures or over debris. Examples of such environments
are shown in Figure~\ref{fig:examples}.

Typically, current approaches solve this problem hierarchically: a top-level
planner decomposes the complete task into smaller tasks and then runs a
different planner, specialized for the each task, in isolation. Once plans are
computed for each task, the top-level planner figures out how the robot will
transition from one task to the next. This can be done with yet another
specialized planner, or prescripted behaviors.
The results of the Darpa Robotics Challenge, as demonstrated in
\cite{lee2014drc}, \cite{atkeson2015drc}, \cite{hauser2014drc}, show the
ubiquity of specialized task planners and behaviors.
This common approach has shown to
be brittle, as each task planner is constrained to satisfy the requirements of
the original task decomposition, and must satisfy strict endpoint constraints to
ensure that the transitions between tasks are feasible. In the worst case, where
the top- level planner has chosen an incorrect decomposition, one of the
planners may be unable to generate a solution at all or a transition between
tasks is infeasible.

The approach presented here builds upon the notion of adaptive dimensionality.
Rather than always search through a high-dimensional state space, adaptive
dimensionality automatically figures out what dimensions are relevant in each
region of the state space. This is tremendously beneficial to planning for
humanoid mobility as there is lots of redundant motion that makes up the various
modes of locomotion available to them. This paper presents how adaptive
dimensionality can be applied to humanoid mobility, describes an implementation
that yields real-time execution by interleaving planning and execution, and
presents experimental results showing the practicality of the approach.


\section{Related Work}

Much work in humanoid mobility planning is focused on solving specific sub-problems
of mobility. Examples of navigation planning using footsteps are shown in
\cite{kuffner2005motion}, \cite{hornung2012anytime}. These techniques plan in a
low-dimensional space representing feasible footstep actions. They may rely on a
controller to produce feasible joint trajectories or the
planner may generate these trajectories online to ensure footstep validity.
Example whole-body planning
techniques have been explored in \cite{burget2013wholebody},
\cite{athar2016whole}, \cite{dalibard2009task}. These approaches are generally
intended for object interaction tasks, and don't consider incorporating
locomotion. Some example techniques specifically for climbing ladders are presented in
\cite{zhang2013ladder} and \cite{kanazawa2015ladder}.

Relatively less work has been done for humanoid robots on adaptively
reasoning about the relevant dimensionality of the problem during the search
process. Some examples of adaptive reasoning include
\cite{vahrenkamp2008adaptive} and \cite{park2014high}. Both works decompose the
robot into appropriate subsystems based on kinematics, increase the
dimensionality. The first RRT-based approach adaptively adds subsystems
as the search gets closer to the goal. The second, optimization-based
approach, plans iteratively, incorporating more descendant subsystems until a
valid path is found. These approaches both iteratively increase the dimensionality of the
entire search space, whereas our approach only increases dimensionality of the search
space in regions where high-dimensional planning is required.

\section{Planning Framework}

Our application to multi-modal humanoid navigation targets the humanoid robot
shown in Figure~\ref{fig:rep_dofs}. The robot has 4 symmetric limbs,
each with 7 degrees of freedom, and additional joints for reorienting the
attached sensors. At the end of each limb, is a dual-purpose end effector
designed with a flat surface, both for walking and support, and a hook for
latching onto cylindrical-shaped objects in the environment, such as ladder
rungs and handrails. In this work, we explicitly plan for all of the joint
variables of the limbs. The pose of the robot provides an additional 6 degrees
of freedom for its global position and orientation. Together these degrees of
freedom define a 34-dimensional search space.

We are also provided with a set of specialized controllers for performing
various locomotion tasks. Currently, the robot is equipped with controllers for
bipedal walking, crawling, and climbing ladders. Additionally, we are able to
directly control each of the joint actuators to execute raw full-body paths.
While it is possible to plan paths consisting of only raw joint motion, we are
able to leverage the existence of the controllers both to improve planning
efficiency and generate plans that can be executed more robustly on the actual
robot.

We represent the planning problem as a search over a finite, discrete search
space. The search space consists of a discrete state space $S$, and a set of
transitions $T = \{(s_i, s_j)|s_i, s_j \in S\}$. Each pair $(s_i, s_j) \in T$
represents a feasible transition between two states. Each transition is
associated with a scalar cost, $c(s_i, s_j) > 0$. We use the notation $\pi(s_i, s_j)$
to denote a path from state $s_i$ to state $s_j$, and $\pi^{*}(s_i, s_j)$ to
denote an optimal, least-cost, path. This search space defines a graph $G$, with
vertex set $S$ and edge set $T$. The goal of the planner is to find a path in
$G$ from a given start state $s_s$ to a goal state $s_g \in S_G$, where $S_G
\subset S$.

To improve the efficiency of the search through this high-dimensional space, and
to ease integration of specialized controllers, our algorithm takes much of its
inspiration from the Planning with Adaptive Dimensionality framework presented
in \cite{gochev2011path} and \cite{gochev2012planning}. The framework for
planning with adaptive dimensionality makes the observation that, in many areas
of the search space, it is often not necessary to reason about the high
dimensionality of the state space, as many of the resulting paths have a low-dimensional
structure.

Section \ref{pad-section} will begin with a brief overview of the framework for
planning with adaptive dimensionality. Section \ref{mrpad-section} will describe
extensions to the planning with adaptive dimensionality framework to enable
planning with multiple low-dimensional planning representations simultaneously.
Section \ref{mrmha-section} will describe the details of the search algorithm
used during the first phase of a single search iteration, with an emphasis on
the use of multi-heuristic search. Section \ref{egraphs-section} will describe
the details of the search algorithm used during the second phase of a single
search iteration, with an emphasis on incorporating user demonstrations to
accelerate planning similar transitions in the high-dimensional space. Section
\ref{pne-section} will describe how the search can be run in a resumable fashion
to enable interleaving of planning and execution. Section \ref{ctrl-section}
will present a brief overview of the control architecture on the robotic
platform, and specifically how plans are delivered to the appropriate
controllers during execution. Section \ref{experiments-section} will list the
results of sample runs of the planner on targeted test environments.

\section{Planning with Adaptive Dimensionality} \label{pad-section}

This section provides a brief overview of the framework for planning with
adaptive dimensionality. For detailed analysis of the adaptive dimensionality
framework and additional applications, see \cite{gochev2011path}.

For a complete planning solution, a search often needs to reason over a high-dimensional
state space. However, we expect that large portions of a complete
plan will exhibit a low-dimensional structure. For example, part of a humanoid
mobility task might include large segments of bipedal walking. In these
scenarios, it suffices to plan only for the footstep locations, and we reserve
planning in the high-dimensional space for verifying that each footstep is
feasible. The portions of the plan requiring high-dimensional reasoning are
infrequent compared to portions that can be solved in this manner.

The planning with adaptive dimensionality framework leverages this low-dimensional
structure by iteratively constructing a hybrid search space,
composed primarily of low-dimensional states and transitions, and introducing
high-dimensional states and transitions where necessary to ensure feasibility
of the resulting path.

\subsection{Graph Structure}

The adaptive dimensionality framework considers two state spaces: the original
high degree-of-freedom state space that represents valid configurations of the
robot, and a projection of the original state space to a low-dimensional
representation, respectively labeled $S^{hd}$ and $S^{ld}$.
A many-to-one mapping defined by
\begin{equation*}
\lambda: S^{hd} \rightarrow S^{ld}
\end{equation*}
represents the projection from the high-dimensional space to the low-dimensional
space. The inverse, one-to-many, mapping defined by
\begin{equation*}
\lambda^{-1}(s_{ld}) = \{s \in S^{hd} | \lambda(s) = s_{ld}\}
\end{equation*}
represents the projection from a state in the low-dimensional space back to a
subset of states in the high-dimensional space.

Both the high-dimensional and low-dimensional space can have its own set
of transitions, $T^{hd}$ and $T^{ld}$ respectively. However, to guarantee
completeness and bounded suboptimality, the following constraint is required:
\begin{equation}
c(\pi^{*}(s_i,s_j)) \geq c(\pi^{*}(\lambda(s_i),\lambda(s_j))), \forall s_i, s_j \in S^{hd}
\end{equation}
That is, the cost of the optimal path between any two states in the high-dimensional
space must be at least the cost of the optimal path between their
projections in the low-dimensional space.

The notation $G^{hd}$ and $G^{ld}$ represent the corresponding high-dimensional
and low-dimensional graphs defined as $(S^{hd}, T^{hd})$ and $(S^{ld}, T^{ld})$,
respectively.

\subsection{Search Algorithm}

Rather than search for a path in the original high-dimensional search space,
$G^{hd}$, the adaptive dimensionality search algorithm prefers to search as much
as possible in the low-dimensional search space, $G^{ld}$. To accomplish this,
the search iteratively constructs a new hybrid search space $G^{ad}$, composed
of an adaptive state space $S^{ad}$, and transition set $T^{ad}$. This new
search space is composed primarily of states and transitions from $G^{ld}$ and
is expanded to include regions of states and transitions from $G^{hd}$ as
necessary.

Initially, the adaptive search space $G^{ad}$ includes all of $G^{ld}$. When a
region of high-dimensional states is introduced, $G^{ad}$ is updated so that
low-dimensional states $s$ that fall within the high-dimensional region are
replaced by their high-dimensional equivalents in $\lambda^{-1}(s)$.

To be able to search this hybrid space, we must define a transition set that
includes transitions between states from $S^{ld}$ and $S^{hd}$. The transition
set for the adaptive search space is defined as follows. For a state $s \in S^{ad}$,
\begin{itemize}

\item
    If $s \in S^{hd}$ then
        for all transitions $(s, s') \in T^{hd}$,
            if $s' \in S^{ad}$ then
                $(s, s') \in T^{ad}$
            otherwise
                $(s, \lambda(s')) \in T^{ad}$

\item
    If $s \in S^{ld}$ then
        for all transitions $(s, s') \in T^{ld}$,
            if $s' \in S^{ad}$ then
                $(s, s') \in T^{ad}$.
        Additionally, for all transitions $(s_{hd}, s_{hd}') \in T^{hd}$,
                where $s_{hd} \in \lambda^{-1}(s)$,
            if $s_{hd}' \in S^{ad}$ then
                $(s, s_{hd}') \in T^{ad}$

\end{itemize}
This transition set includes transitions between low- and high-dimensional
states, and only includes transitions to states in the adaptive state space
$S^{ad}$. Notice that expanding or adding a new high-dimensional produces a new
instance of $G^{ad}$.

The adaptive search algorithm begins by finding a path, $\pi_{ad}$, from the
start to the goal in the current instance of $G^{ad}$. This path is allowed to
contain states of differing dimensionalities, and so may not be executable. If
no path is found during this phase, then no path exists from the start to the
goal, and the search terminates. To construct an executable path from
$\pi_{ad}$, another search is conducted within a tunnel surrounding $\pi_{ad}$.
We define a tunnel $\tau$ of radius $w$ around an adaptively-dimensional path
$\pi_{ad}$ as follows: $\tau$ is a subgraph of $G^{hd}$. A high-dimensional
state $s \in \tau$ if there exists a state $s_{ad} \in \pi_{ad}$ such that the
distance from $\lambda(s)$ to $s_{ad}$ (or $\lambda(s_{ad})$ if $s_{ad} \in S^{hd}$)
is no larger than $w$ for some pre-defined distance metric on $S^{ld}$. All
transitions $(s, s') \in T^{hd}$ are included such that $s, s' \in \tau$.

If the search fails to find a path from the start to the goal within $\tau$,
high-dimensional regions are introduced where the search became stuck, and the
adaptive search begins a new iteration on a newly constructed instance of
$G^{ad}$. See \cite{gochev2011path} for details on how to identify locations
to place high-dimensional regions.

\section{Adaptive Dimensionality with Multiple Low-Dimensional Representations} \label{mrpad-section}


In the domain of humanoid mobility planning, several useful low-dimensional
representations are available. For our application, the humanoid robot is
expected to utilize the available controllers for optimized bipedal walking,
crawling, and ladder climbing. Each of these controllers has a natural low-dimensional
representation. For crawling, the controller requires a 4-dimensional pose,
$(x, y, z, \theta)$, of the robot.  For bipedal walking, the controller requires
paths that specify the 4-dimensional pose, $(x, y, z, \theta)$, of each foot.
Finally, for ladder climbing, the controller requires only the 6-dimensional
pose, $(x, y, z, \alpha, \beta, \gamma)$, for each of the four end-effectors.
To be able to plan solutions that incorporate all of these representations, we
need the ability to combine into a single search space.

\begin{figure}
    \centering
    \includegraphics[width=\linewidth]{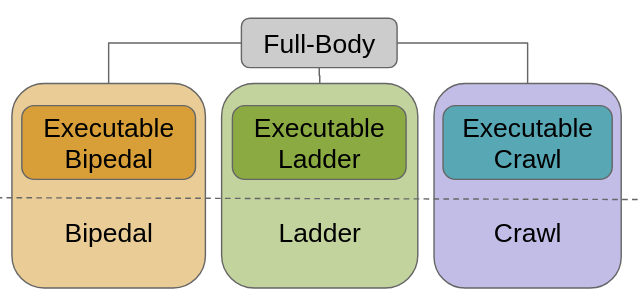}
    \caption{Relationship between Representations}
    \label{fig:rep_relations}
\end{figure}

Our approach maintains the separation between the high-dimensional and low-dimensional
search spaces and their ability to define their own transition sets. Given $n$
low-dimensional representations, we define low-dimensional discrete state spaces
$S^1, S^2, \ldots, S^n$, and their corresponding transition sets, $T^1, T^2, \ldots, T^n$.

The mappings from the high-dimensional space to each low-dimensional space
remain largely unchanged as well. For the $i$'th low-dimensional
representation, a mapping defined by
\begin{equation*}
\lambda_i: S^{hd} \rightarrow S^i
\end{equation*}
represents the mapping from states in $S^{hd}$ to states in $S^i$. Correspondingly,
the inverse functions
\begin{equation*}
\lambda_i^{-1}(s_i) = \{s \in S^{hd} | \lambda(s) = s_i\}, \forall i \in 1..n
\end{equation*}
represent the mapping from states in $S^i$ to subsets of $S^{hd}$.

Additionally, we define functions
\begin{equation*}
\lambda_{i,j}(s_i) = \{s \in S_j | \exists s_{hd} \in \lambda^{-1}(s_i) [ \lambda(s_{hd}) = s ] \}
\end{equation*}
to represent the mappings between states of low-dimensional representations.
These mappings may be one-to-one or one-to-many depending on the dimensionality of
the target representation.

The construction of $S^{ad}$, and its graph representation, $G^{ad}$, follows
from its construction in the adaptive dimensionality framework. The initial
instance of $G^{ad}$ is the union of the search spaces $G^i = (S^i, T^i)$ for
each of the low-dimensional representations. The transition set, $T^{ad}$ is
extended to include projections from the high-dimensional representation to each
low-dimensional representation. Additionally, $T^{ad}$ also contains transitions
that allow the search to effectively switch between low-dimensional
representations. The complete transition set is defined as follows.
For a state $s \in S^{ad}$,
\begin{itemize}

\item
    If $s \in S^{hd}$ then,
        for all high-dimensional transitions $(s, s') \in T^{hd}$,
            if $s' \in S^{ad}$ then
                $(s, s') \in T^{ad}$,
            otherwise
                $(s, \lambda_i(s')) \in T^{ad}$ for each low-dimensional representation $S^i$

\item
    If $s \in S^i$ then,
        for all low-dimensional transitions $(s, s') \in T^i$,
            if $s' \in S^{ad}$ then
                $(s, s') \in T^{ad}$.
        Additionally, for each representation $S^j \in \{S^k | k \in 1..n, k \neq i \} \cup S^{hd}$, 
            for all transitions $(s_j, s_j') \in T^j$,
                    where $s_j \in \lambda_{i,j}(s)$,
                if $s_j' \in S^{ad}$ then
                    $(s, s_j') \in T^{ad}$

\end{itemize}

Thus far, we have only described how to incorporate multiple low-dimensional
representations into the adaptive dimensionality framework. This indeed speeds
up the search for finding a high-dimensional path, but we also desire to
explicitly reason about the controller capabilities, to avoid high-dimensional
planning wherever possible. Recall that during the second phase of each search
iteration, the algorithm searches for a completely high-dimensional path, within
the tunnel $\tau$. To relieve the search of needing to perform high-dimensional
planning, we extend the transition set of the high-dimensional representation to
include all of the transitions that correspond to actions from the low-dimensional
representations that are directly executable by an available
controller.

\section{Low-Dimensional Representations for Humanoid Mobility} \label{reps}

\begin{figure}
    \centering
    \begin{subfigure}{0.30\linewidth}
        \includegraphics[width=\linewidth]{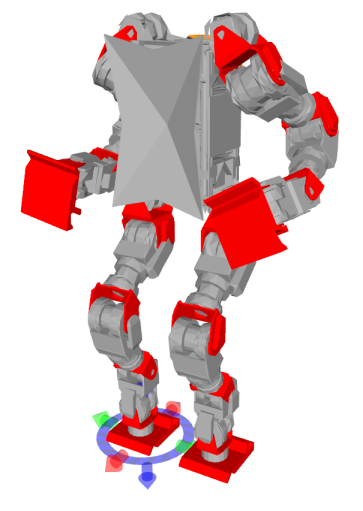}
        \caption{}
        \label{fig:dofs_biped}
    \end{subfigure}
    \begin{subfigure}{0.30\linewidth}
        \includegraphics[width=\linewidth]{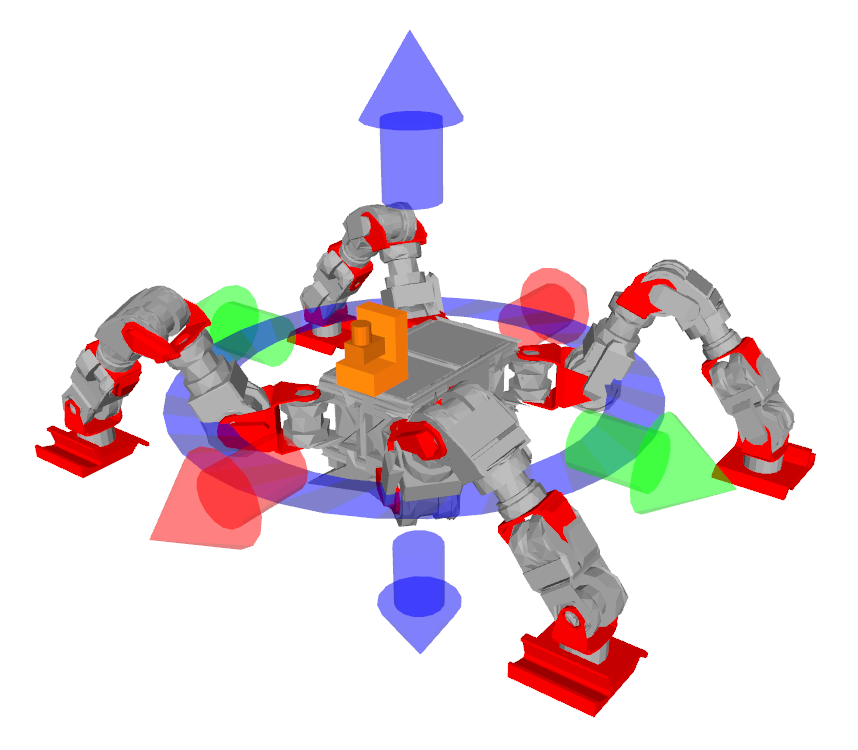}
        \caption{}
        \label{fig:dofs_crawl}
    \end{subfigure}
    \begin{subfigure}{0.30\linewidth}
        \includegraphics[width=\linewidth]{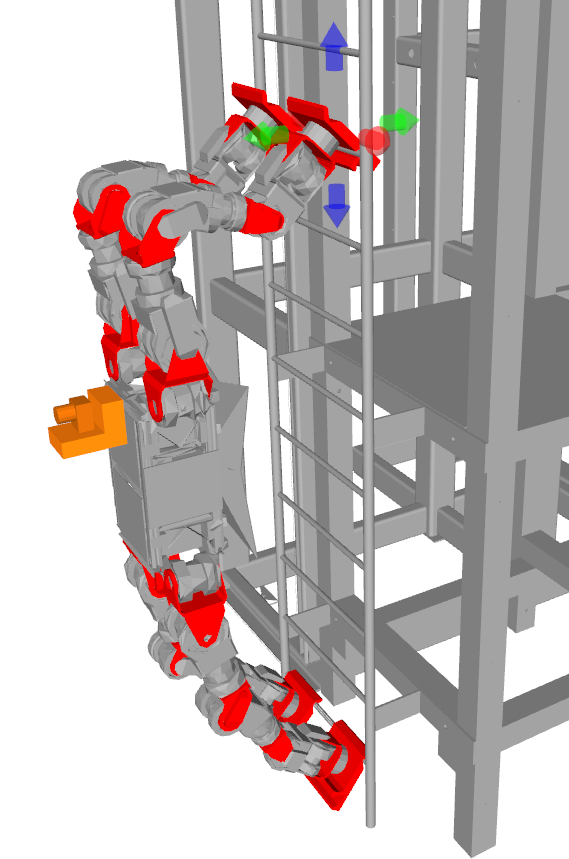}
        \caption{}
        \label{fig:dofs_ladder}
    \end{subfigure}
    \caption{Humanoid Developed by Mitsubishi Heavy Industries, Ltd. and its Low-Dimensional Representations}
    \label{fig:rep_dofs}
\end{figure}

In this domain, the high-dimensional state space represents all the controllable
degrees of freedom of the robot. Each low-dimensional state space represents
one of the available modes of locomotion. These low-dimensional representations
are depicted in Figure~\ref{fig:rep_dofs}.

A state vector for the full-body state space contains a single discrete variable
for each actuated joint, plus 6 variables for the pose of the robot. Combined,
a state vector for the full-body state space is represented as
\begin{equation*}
(pose_x, pose_y, pose_z, pose_\phi, pose_\theta, pose_\psi, j_1, j_2, \ldots, j_{28})
\end{equation*}
Each discrete variable corresponds to a range of continuous values, obtained by
simple discretization functions. The discretization resolution was chosen as
\SI{1}{cm} for all translational variables, and as \SI{5}{\degree} for all
rotational variables. The action space is built from several types of motion
primitives. The first type directly moves each joint of the robot individually
by some small delta. We chose simple motion primitives that varied each joint by
the state space discretization of \SI{+-5}{\degree}. The second type of motion
primitive uses an inverse kinematics solver to move the position of one of the
end effectors by a small positional delta. We chose simple primitives to move an
individual end effector by \SI{+-2.5}{cm} in x, y, or z. Finally, we allow
full-body IK motions for the root of the robot. The root is allowed to move in
x, y, or z by \SI{+-5}{cm} or yaw by \ang{12.25}, \ang{22.5}, or \ang{45}. Since
there were no controllable degrees of freedom between the root of the robot and
the base of each limb, our full-body body IK solver simply runs an isolated IK
solver for each of the limbs that are currently supporting the weight of the
robot. The last type of motion primitive is an adaptive motion primitive that
computes the motion, on-the-fly, that achieves a selected target for one of the
limbs. These targets are selected according to nearby support surfaces, such as
the ground or handrails.

The representation for bipedal walking contains state vectors
describing the 4-dimensional $(x, y, z, \theta)$ poses of each of the foot, plus
one extra variable for restricting the gait of the robot to a left-right
alternating scheme. The combined state vector is represented as
\begin{equation*}
(x_l, y_l, z_l, \theta_l, x_r, y_r, z_r, \theta_r, pivot)
\end{equation*}
where $pivot$ is $left$ or $right$ to denote the pivot foot. We restrict the
actions allowed in the bipedal state representation to a fixed set of target
poses, offset from the current pivot foot. Application of one of these actions
places the opposite, active foot with respect to the pivot foot. We included
$15$ total primitives: $2$ of these allow reorienting of the feet to produce
turning motions, and the remaining $13$ move the feet forward at varying
distances from \SI{2}{cm} to \SI{24}{cm} to trade off between state space
coverage and allowing the search to quickly explore long distances via
forward walking.

The representation for crawling contains state vectors describing a
four-dimensional $(x, y, z, \theta)$ pose for the center-of-mass of the robot.
We include a simple set of actions that allow the robot to move directly forward
or backward by $10cm$, and to turn-in-place by \ang{45}, to mimic the capabilities
of the controller.

The representation for ladder climbing contains state vectors
describing a six-dimensional $(x, y, z)$ positions for each of the robot's end
effectors. Each action moves all four end effectors from their current positions
to positions on the ladder rung directly above or below the currently held rung,
as indicated by the current end effector positions.

The projection functions, $\lambda_1, \ldots, \lambda_n$
from the full-body representation to each of these low-dimensional
representations require solving forward kinematics for the given full-body
state. The inverse projections from each low-dimensional representation to the
high-dimensional representation require expensive inverse kinematics queries to
determine valid configurations for the robot. To accelerate this process, we
precompute a small set of nominal joint configurations for the full-body state,
and perform small searches for a nearby valid configuration. Conveniently, these
nominal joint configurations also correspond to the joint
configurations we can expect the robot to achieve after execution of
one of the controllers.

Notice that because of the restrictions on the available actions for each state
representation, the planner may be unable to discover relevant projections
between the state representations. To aid the planner, we append special
transitions, whose resulting successor states project cleanly
to another state representation. For example, a successor state for bipedal is
generated such that the feet end up on a ladder rung, so that the resulting
projection is relevant to the ladder representation. We designed similar
transitions from bipedal to crawling and from crawling to ladder.
These special successor states are only for projections to other representations,
and may not produce successor states of their own.

By design, the available actions in the crawling and ladder representations are
always directly executable by an existing controller. However, the bipedal
representation contains actions that are not directly executable. For example,
the bipedal representation is allowed to make footsteps that traverse up and
down staircases, while the available controller is only allowed to operate on
even terrain. These non-executable actions are resolved during the second phase
of the search by planning in the high-dimensional space.

\section{MR-MHA*} \label{mrmha-section}

In this section we describe a planning algorithm which is better suited to
planning in multiple representation state-spaces. This algorithm,
called the MultiRep-MultiHeuristic A* (MR-MHA*), is a generalization of the MHA*
algorithm \cite{aine2016multi} that can reason over several different state-space
representations, each of which may have its own heuristics defined.


Multi-Heuristic A* is a search framework that uses multiple
inadmissible heuristics to simultaneously explore the search space, while
preserving guarantees of completeness and suboptimality bounds by using a single
admissible "anchor" heuristic. The algorithm has shown success in complex high-dimensional
planning problems such as mobile manipulation planning for the 12D
PR2 robot, where a naive weighted-A* approach is sensitive to large local
minima. Two variants of MHA* are described in \cite{aine2016multi}. In this work
we simply refer to the shared variant SMHA* as MHA*.


As mentioned earlier, humanoid mobility presents several low-dimensional
representations like bipedal walking, ladder climbing etc. It would be
preferable for the search to explore along both the ladder and stairway at the
same time to reach the goal. Also since these low-dimensional representations
are complex enough and fundamentally different, we might need different
heuristics to explore each representation. This is the motivation for
generalizing MHA* to a multi-representation setting.
Note that MR-MHA* simplifies to vanilla MHA* algorithm for
planning with a single representation. The full-body tracking phase of planning
for humanoid mobility with adaptive dimensionality as in our case is one such
example.

\begin{algorithm}
	\scriptsize{
	\caption{MR-MHA*}
	\label{alg:mrmha_algo}
	\begin{algorithmic}[1]
	\Procedure{InitializeHeuristicList}{}
		\For {$d = 1 \ to \ max\_dim$}\
			\State{$heuristic\_list[d].anchor = h_0$}
			\For{$i = 1\ to\ n$}
				\If{$h_i$ is enabled for dim $d$}
					\State{$heuristic\_list[d].inadm.append(h_i)$}
				\EndIf
			\EndFor
		\EndFor
	\EndProcedure
	\Procedure{key}{$s$, $i$}
		\State{\textbf{return} $g(s) + w_1 \times h_i(s)$}
	\EndProcedure
	\Procedure{expand}{s}
		\State{Remove $s$ from $OPEN_i$ $\forall$ $i$ in $heuristic\_list[s \rightarrow dim]$}
		\For{each s' in Succ(s)}
			\If{s' was never visited}
				\State{$g(s') = \infty;bp(s') = null $}
			\EndIf
			\If{$g(s') > g(s) + c(s,s')$}
				\State{$g(s) = g(s') + c(s,s'); bp(s') = s$}
				\If{$s'$ has not been expanded in the $anchor$ search}
					\State{insert/update $s'$ in $OPEN_0$ with $key(s',0)$}
					\If{$s'$ has not been expanded in any $inadmissible$ search}
						\For{$i\ in\  heuristic\_list[s' \rightarrow dim].inadm$}
							\If{$key(s',i) \leq w_2 \times key(s',0)$}
								\State{insert/update $s'$ in $OPEN_i$ with $key(s',i)$}
							\EndIf
						\EndFor
					\EndIf
				\EndIf
			\EndIf
		\EndFor
	\EndProcedure
	\Procedure{MR-MHA*}{}
		\State{$g(s_{goal}) = \infty; bp(s_{start}) = bp(s_{goal}) = null$}
		\State{$g(s_{start}) = 0$}
		\State{InitializeHeuristicList()}
		\For{$i = 0\ to\ n$}
			\State{$OPEN_i = \emptyset$}
			\If{$i in heuristic\_list[s_{start} \rightarrow dim]$}
				\State{insert $s_{start}$ into $OPEN_i$ with $key(s_{start},i)$ as priority}
			\EndIf
		\EndFor
		\While{$OPEN_0$ not empty}
			\For{$i = 1\ to\ n$}
				\If{$OPEN_i.MinKey() \leq w_2 \times OPEN_0.MinKey()$}
					\If{$g(s_{goal}) \leq OPEN_i.Minkey()$}
						\State{terminate and return path pointed by $bp(s_{goal})$}
					\EndIf
					\State{$s = OPEN_i.Top()$}
					\State{expand(s)}
				\Else
					\If{$g(s_{goal}) \leq OPEN_0.Minkey()$}
						\State{terminate and return path pointed by $bp(s_{goal})$}
					\EndIf
					\State{$s = OPEN_0.Top()$}
					\State{expand(s)}
				\EndIf
			\EndFor
		\EndWhile
	\EndProcedure
	\end{algorithmic}
	}
\end{algorithm}

\subsection{Algorithmic Details}

\subsubsection{Heuristic Lists}


Following MHA*, we have a single admissible heuristic across all low-dimensional
representations to satisfy suboptimality bounds on the solution obtained from
the low-dimensional search space. Of all $n$ possibly inadmissible heuristics
that are available to the search, a subset of them is available to each low-dimensional
representation, depending on whether a particular heuristic is
enabled for that representation. This splitting of heuristics between
representations allows the search to explore simultaneously across
representations. The high-dimensional representation has its own anchor and
set of inadmissible heuristics, as in vanilla MHA*, since the searches for
the two adaptive planning phases are independent. This is defined in
the $InitializeHeuristicLists()$ method in lines $1-6$ in the algorithm.


\subsubsection{Successor Generation}

In MHA*, whenever a state is expanded, its successors are inserted into all
inadmissible heuristic queues that are available to the search provided it has
not been expanded from either the anchor or any of the inadmissible searches. This
enables MHA* to effectively share paths between different heuristics that can
help the search in different parts of the state-space.

However, when a state is expanded in MR-MHA*, the representation dimension of each
successor is extracted, and accordingly are only inserted in heuristic queues
which are available to that particular representation as defined in the
heuristic lists. This allows MR-MHA* to effectively share paths within each
representation without unnecessarily expanding states from irrelevant
heuristic queues. This is shown in line $19$ in the algorithm.

\subsection{Implementation Details}

Here we summarize the heuristics used for each state-space representation in our
framework. As mentioned before, we perform full-body planning only in parts of
the state-space where a controller is not executable. In our experimental setup,
this only corresponds to the humanoid stepping on stairs. Hence, the heuristics
we list for the full-body representation aid the search in humanoid stepping
motion only.

A common approach to designing a heuristic function for a given state space is
to first project it to a low-dimensional representation, and use the result of a
search in the low-dimensional space as the heuristic value for a corresponding
high-dimensional state. We designed several 3D grid searches with cost functions
tuned for producing meaningful heuristic values, and computed those values
online using a Dijkstra search from the goal to the state whose heuristic we
are computing.

\textbf{Bipedal Representation}

\begin{enumerate}
\item { Sum of grid search distances from both feet to the goal }
\item { Sum of grid search distances from both feet to the goal, with penalties for stepping close to the edges of staircase steps, to encourage alignment with the staircase direction }
\item { Sum of grid search distances from both feet to the goal, with penalties for using the ladder, to encourage staircase usage }
\end{enumerate}

\textbf{Ladder Representation}

\begin{enumerate}
\item { A constant 0, to expand states in order of increasing g-values }
\end{enumerate}

\textbf{Crawl Representation}

\begin{enumerate}
\item { Grid search distance from the COM of the robot to the goal }
\end{enumerate}

\textbf{Full-Body Representation (Stepping)}

\begin{figure}[H]
\centering
\includegraphics[width = 0.15\textwidth]{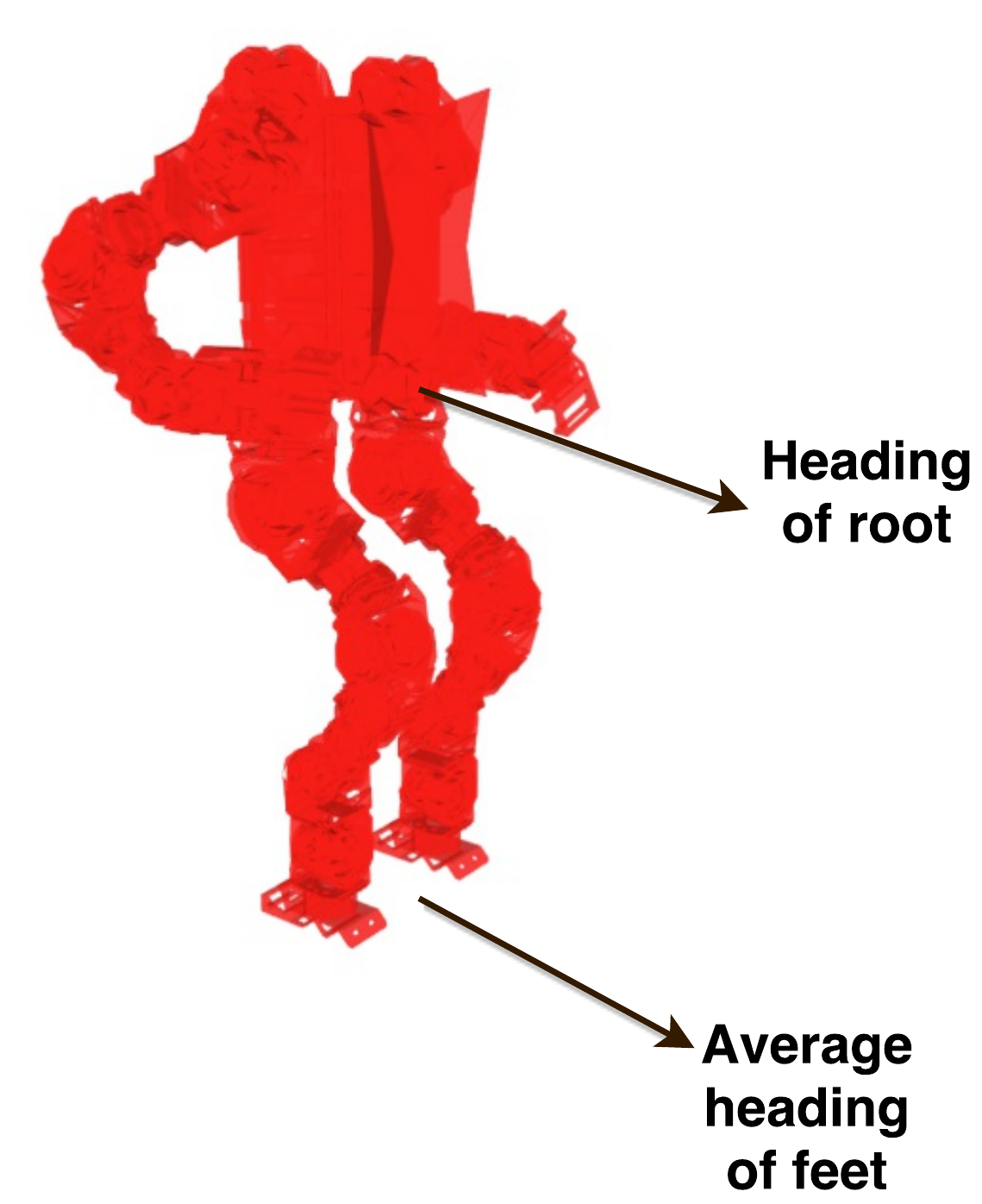}
\includegraphics[width = 0.1\textwidth]{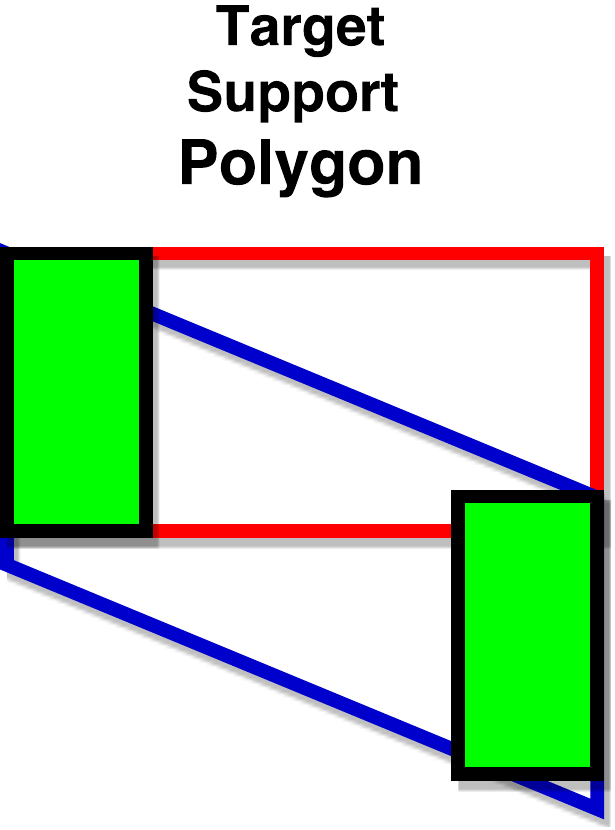}
\includegraphics[width = 0.2\textwidth]{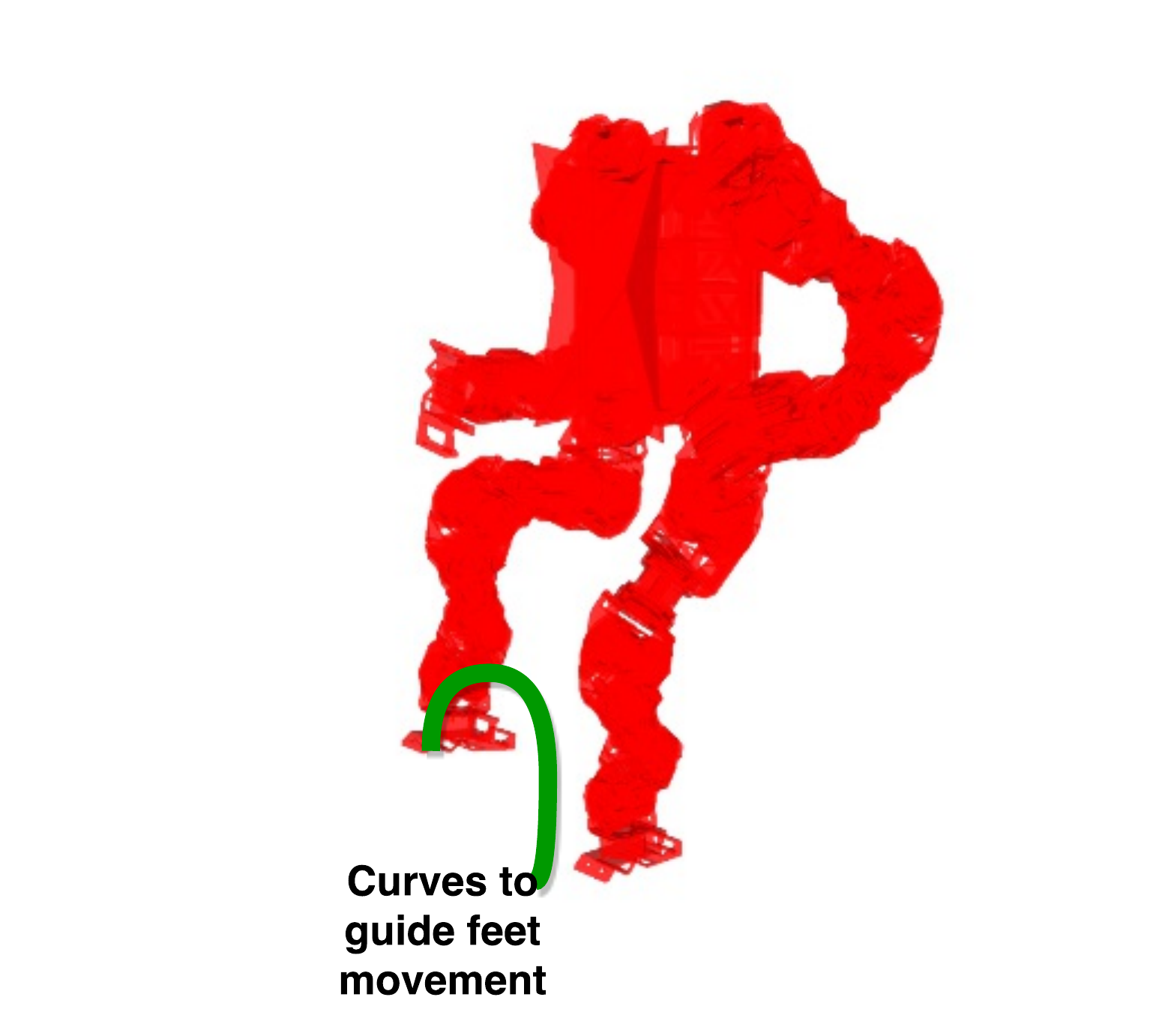}
\caption{Full body representation heuristics.}
\label{fig:highD_heuristics}
\end{figure}

\begin{enumerate}
\item { Grid search distance from the COM of the robot to the goal }
\item { Difference in heading between the root of the robot and the feet of the robot (Fig.\ref{fig:highD_heuristics}) }
\item { Euclidean distance between COM of the current state and that of the target state }
\item { Euclidean distance between active feet of the current state and that of the target state (Fig.\ref{fig:highD_heuristics}) }
\item { Curve to provide guidance for stepping feet movement during the search (Fig.\ref{fig:highD_heuristics}) }
\item { Remaining number of steps on the lower dimensional path that need to be tracked }
\end{enumerate}

\section{Experience Graphs} \label{egraphs-section}

This section details the method used for accelerating planning when the search
must find paths through the high-dimensional space during the second phase of
each search iteration. Many of the transitions on the adaptively-dimensional
path $\pi_{ad}$ are directly executable by one of the available controllers, but
some transitions require planning in the high-dimensional space. The transitions
we focus on in this section are specifically those transitions for the full-body
which move between states of different dimensionalities. For example, during the
first phase, the planner may produce a transition from the bipedal
representation to the ladder representation, which corresponds to the motion for
the robot that mounts the ladder while standing in front it.


To speed up high-dimensional planning in these scenarios, we apply Experience
Graphs, outlined in \cite{phillips2012graphs}. Experience Graphs, or E-Graphs,
provide a way to incorporate prior experience, in our case from user
demonstrations, to guide the search towards reusing paths with a good chance of
leading to the goal. In our domain, these transitions are often similar and we
can leverage previous solutions to generate modified transitions quickly.



\subsection{Heuristic Computation}

As discussed in \cite{phillips2012graphs}, the E-Graphs approach defines the
heuristic value for a state $s_0$ as
\begin{equation*}
h^E(s_0) = \min_{\pi} \sum_{i = 0}^{N-1} \min \{\epsilon^E h^G (s_i, s_{i+1}), c^E(s_i, s_{i+1})\}
\end{equation*}
where $\pi$ is a path $\langle s_0 \ldots s_{n-1} \rangle$, $s_{N-1} = s_{goal}$, and
$\epsilon^{E}$ is a scalar parameter $\geq 1$, which determines the degree to which the
search is encouraged to reuse prior experience. The paths $\pi$ consist of edges
between any two states $s_i$ and $s_{i+1}$ with cost equal to the underlying
heuristic $h^G$, inflated by $\epsilon^E$, and edges from the E-Graph with cost equal to the
actual cost of the transition.

The underlying heuristic used with the E-Graph heuristic is computed by solving
a lower-dimensional problem using dynamic programming. Similar to the heuristics
used during the first phase of the search, we solve several 3D $(x, y, z)$
Dijkstra searches, each from the goal position of one of the end effectors. The
E-Graph transitions, as well as obstacles, are incorporated directly into these
Dijkstra expansions to encourage obstacle avoidance and use of prior experience.
When computing the heuristic value for a given full-body state, we sum up the
contributions from these Dijkstra searches, using the current end effector
positions.

\begin{figure}
	\centering
	\begin{subfigure}{0.24\linewidth}
		\includegraphics[width=\linewidth]{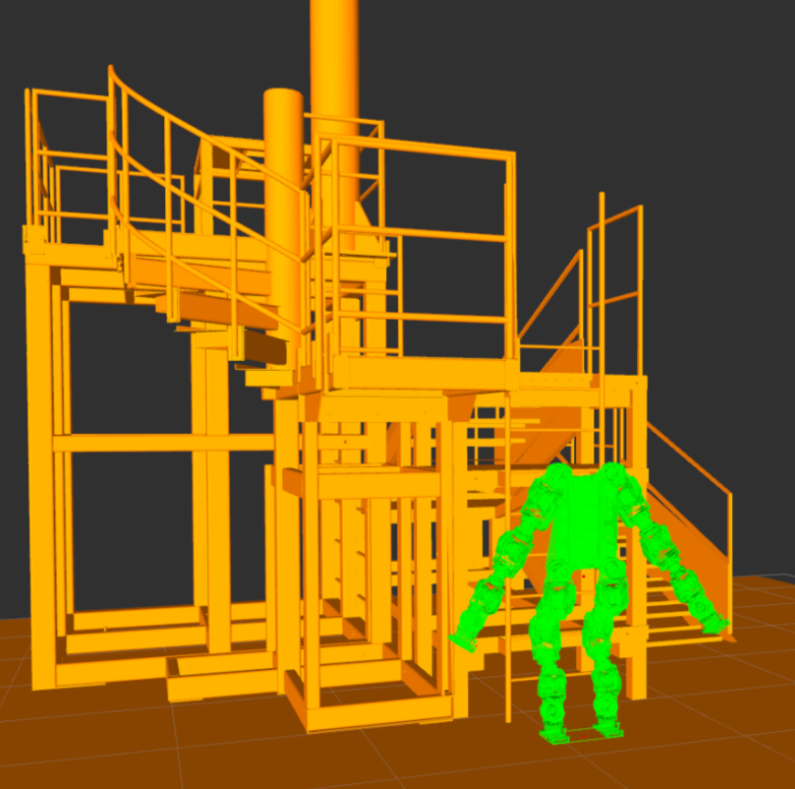}
		\caption{}
		\label{fig:demo1}
	\end{subfigure}
	\begin{subfigure}{0.24\linewidth}
		\includegraphics[width=\linewidth]{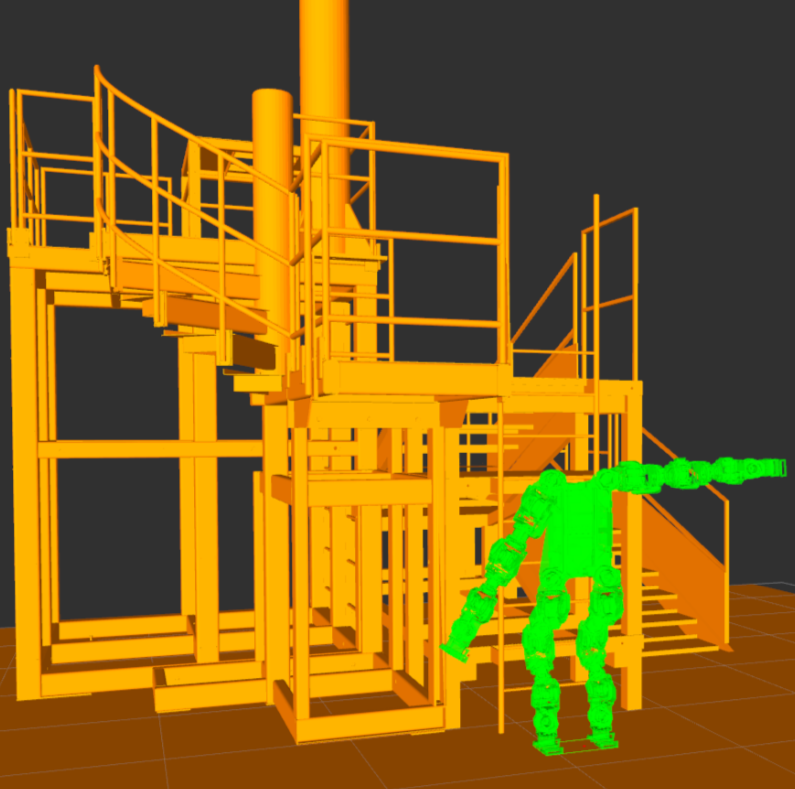}
		\caption{}
		\label{fig:demo2}
	\end{subfigure}
	\begin{subfigure}{0.24\linewidth}
		\includegraphics[width=\linewidth]{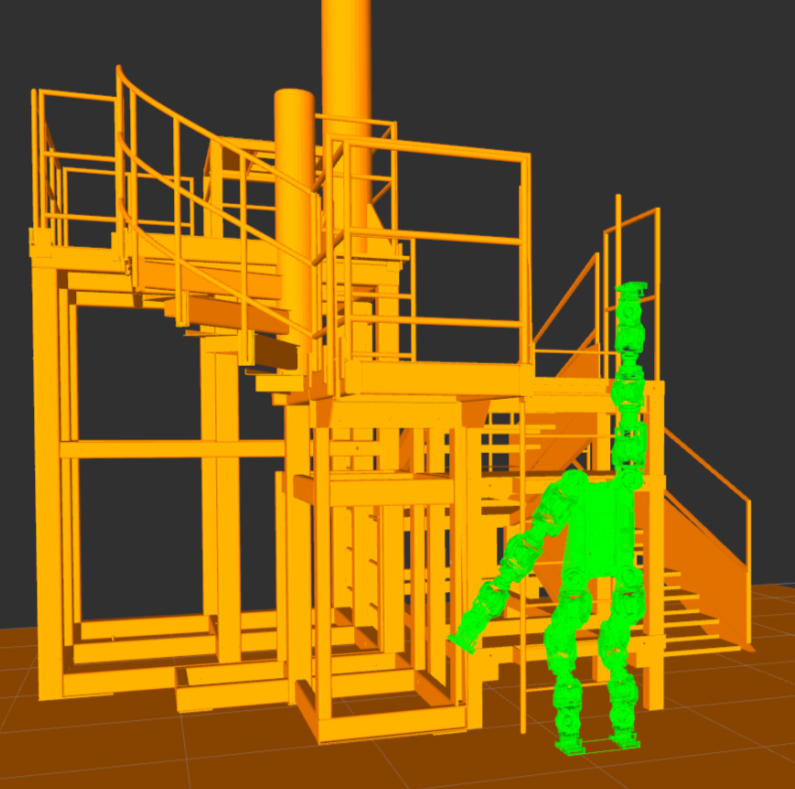}
		\caption{}
		\label{fig:demo3}
	\end{subfigure}
	\begin{subfigure}{0.24\linewidth}
		\includegraphics[width=\linewidth]{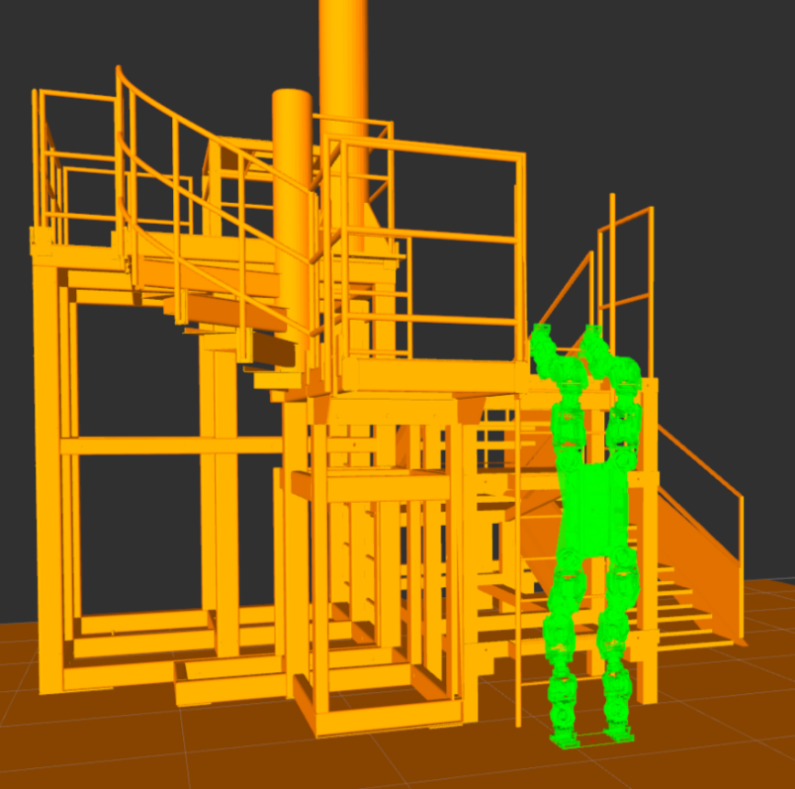}
		\caption{}
		\label{fig:demo4}
	\end{subfigure}
	\begin{subfigure}{0.48\linewidth}
		\includegraphics[width=\linewidth]{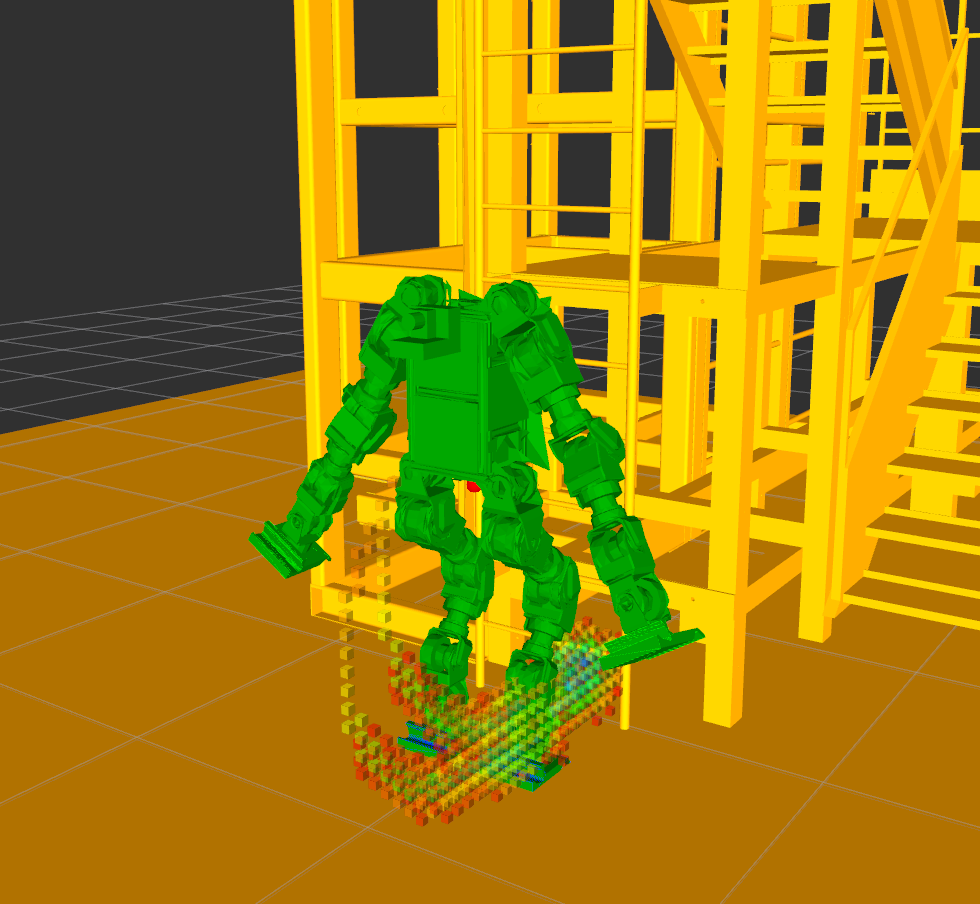}
		\caption{}
		\label{fig:heur}
	\end{subfigure}
	\begin{subfigure}{0.48\linewidth}
		\includegraphics[width=\linewidth]{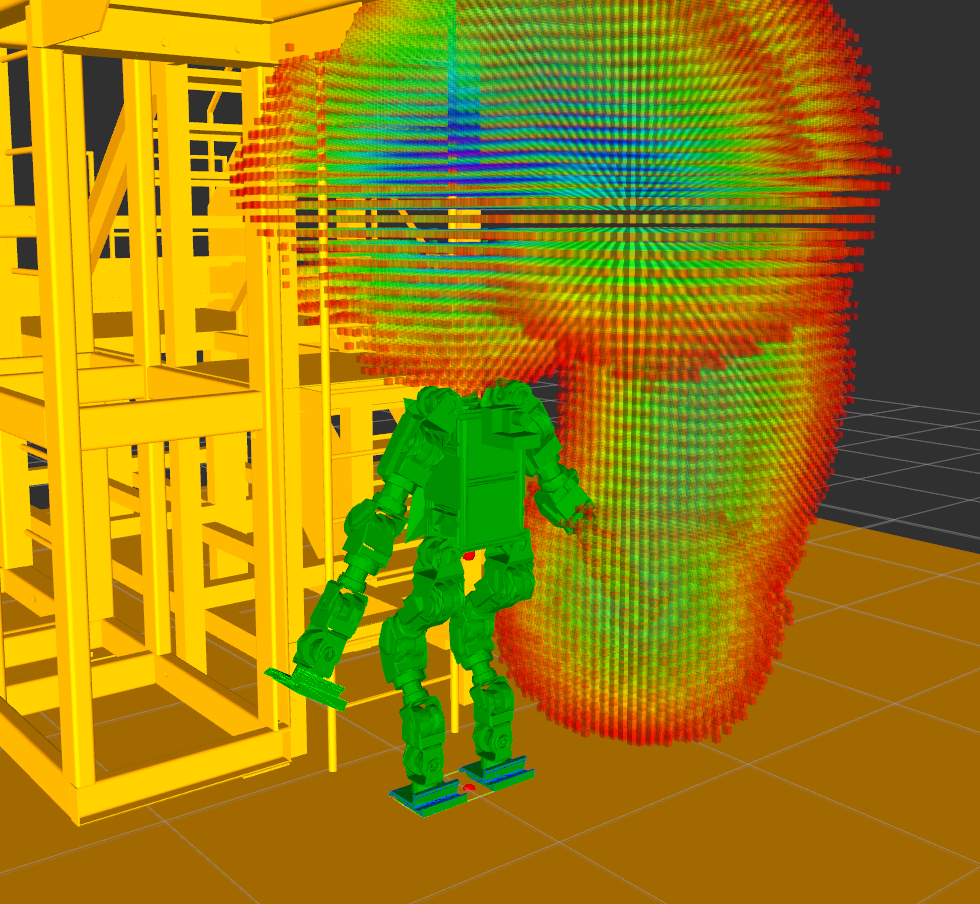}
		\caption{}
		\label{fig:rh_heur}
	\end{subfigure}
	\caption{
		Figure~\ref{fig:heur} and Figure~\ref{fig:rh_heur} illustrate the
		heuristic values for the demonstration in Figure~\ref{fig:demo1} through
		Figure~\ref{fig:demo4}. Cells in blue have lower heuristic values.
	}
	\label{fig:heuristic}
\end{figure}


\subsection{Snap Motions}

In many scenarios, the search can encounter difficulty using the path
demonstrations, even if the path is representative of the expected path the
robot should take. For example, consider a scenario where a user has
demonstrated how the robot should mount a ladder. If, during planning, the
search considers a state where the feet are slightly offset from the
demonstration, it will still have to reason about the motion required to adjust
the feet so that the demonstration can be reused. To alleviate this problem, we
combine the ideas from \cite{phillips2012graphs} and \cite{Cohen:2011} to create
a set of adaptive motion primitives that reach partial states on the E-Graph.
For example in the aforementioned scenario, the adaptive motion primitive will
attempt to match partial waypoints for the arms and the torso, rather than
trying to adjust the feet to match the demonstration completely.

\section{Interleaved Planning and Execution} \label{pne-section}

Owing to the complexity of the planning tasks that we are addressing, the
typical planning times to plan a path all the way from the start to the goal are
significantly high. The key idea here is that instead of waiting for the planner
to generate the entire executable path, we will interleave path planning and
execution. The planner will return partial plan as it runs while the controller
will start executing these plans on the robot in parallel. This idea has been
widely studied by the real-time family of heuristic search algorithms.
For the humanoid domain this approach provides significant speed
ups in the overall planning and execution time because the path execution is
generally slow.

Within the underlying framework of planning with adaptive dimensionality, we
only employ the interleaving scheme in the tracking phase, because the tracking
phase returns a path that is executable by the robot.

\begin{algorithm}
\scriptsize{
\caption{Interleaved Planning and Execution}\label{euclid}
\begin{algorithmic}[1]
\Inputs{lookahead}
\While{tracking time $\not=$ lookahead}
  \State Run Tracking
\EndWhile
\State Reconstruct partial path
\State Send partial path to the controller for execution
\State Reset start state with the tail of the partial path
\State Loop
\end{algorithmic}
}
\end{algorithm}

\section{Controller Framework} \label{ctrl-section}

The path returned by the planner contains both transitions that correspond to
executing an available controller, e.g. walking, crawling, climbing, and those
that correspond to raw full-body joint motion.
As described in Section \ref{mrpad-section}, the low-dimensional transitions are
directly executable by a controller.
It is then the responsibility of the controller to compute full-body
joint trajectories that robustly execute the desired motion. Each individual
controller only accepts paths of their respective waypoint type. To interface
the controllers with the planner, we developed a meta-controller for dispatching
segments of the hybrid path to the correct controller. After a hybrid path is
received from planner, the meta-controller divides it into the individual
segments of the same waypoint type and dispatches them sequentially to the
corresponding controllers. After a segment is executed successfully, the
controller signals the meta-controller to proceed. The meta-controller then
dispatches the next segment to the corresponding controller, until the path is
completed. The meta-controller is also responsible for updating the path, as
additional waypoints are received during interleaved planning and execution.

\section{Experimental Analysis} \label{experiments-section}

\begin{table}
\centering
\begin{tabular}[width=\columnwidth]{
    |r|
    c|c|c|c|
}
\hline
& \multicolumn{2}{c|}{success \%} & \multicolumn{2}{c|}{mean time (s)} \\
\hline
goal & plan & track & plan & track \\
\hline
top & 89.6 & 57.5 & 58.79 & 42.53 \\
mid & 89.6 & 65.7 & 57.67 & 33.61 \\
\hline
\end{tabular}
\caption{}
\label{table:results}
\end{table}

To demonstrate the effectiveness of the multi-heuristic adaptive planning
approach, we tested the planner's ability to plan paths in the sample
environment from Figure~\ref{fig:test_env_2}. The tests consisted of running the
planner from numerous start locations, evenly distributed across (x, y)
locations in the environment and from different start headings, to goal
locations on the mid- and top-level platforms. In all cases, the planner was
given \SI{80}{\second} to find a low-dimensional path, and \SI{180}{\second}
to find the high-dimensional path. The results are shown
Table~\ref{table:results}.

The table lists the results across 231 different start locations, for each goal.
Success rates are shown for both the low-dimensional phase of the search, and
for the high-dimensional of the search. The success rates for the second
"tracking" phase, are normalized with respect to the success of the first phase,
thus the overall success rates for the planner are $51.52\%$ and $58.87\%$,
respectively for the two goals.

The columns containing planning times are mean times for the two phases of the
search individually. The total average planning times are $101.3$ and $91.3$
seconds, respectively. Note that, outside of experiments, the planning is
interleaved with execution of the path on the robot, so the time spent idle is
only limited by the time taken by the search during the low- dimensional
planning phase, plus a small lookahead for the high-dimensional search. The same
is also true for the success rate, where the search is only limited by the speed
of execution.

\section{Conclusion}

In this work, we have presented an approach to planning for multi-modal humanoid
mobility using a single search algorithm. This approach is able to
simultaneously reason about all the capabilities of the robot, incorporate the
capabilities of available controllers, and automatically discover the
transitions for switching between modes of locomotion. The resulting planner
brings together planning with an adaptively-dimensional search space, using
multiple heuristics, and incorporating user demonstrations to concentrate search
efforts where they're most needed.
In future work we hope to incorporate planning
for more complex interaction with the environment and address the robustness
of scenarios requiring high-dimensional planning.

%

\bibliographystyle{IEEEtran}
\bibliography{references}

\end{document}